\documentclass[letterpaper, 10 pt, conference]{ieeeconf}  
\usepackage[T1]{fontenc}

\usepackage{algorithm}
\usepackage{algpseudocode} 

\usepackage[table]{xcolor}
\usepackage{upgreek}
\usepackage{multicol}
\usepackage{dsfont,setspace}
\usepackage{amsmath,graphicx}
\usepackage{float}
\usepackage{amssymb}
\usepackage{amsmath}
\usepackage[normalem]{ulem}
\usepackage[thinc]{esdiff}

\usepackage{lineno}
\usepackage{epstopdf}
\usepackage{authblk}
\usepackage{dblfloatfix}
\usepackage{tabularx}
\usepackage{soul}

\usepackage[
  separate-uncertainty = true,
  multi-part-units = repeat
]{siunitx}

\usepackage{subcaption}
\usepackage[colorlinks]{hyperref}
\hypersetup{
     colorlinks,
     linkcolor=black,
     citecolor=black,
     filecolor=black,
     urlcolor=blue,
 }
\usepackage{cleveref}
\usepackage{textcomp}
\usepackage{gensymb}
\usepackage{tikz}
\usepackage{pifont}
\usepackage{makecell}
\usepackage{svg}

\newcommand{\cmark}{\ding{51}}%
\newcommand{\xmark}{\ding{55}}%

\newcommand{\grey}[1]{
{\color{gray}{#1}}
}

\newcommand{\red}[1]{
{\color{red}{#1}}
}

\newcommand{\green}[1]{
{\color{teal}{#1}}
}

\IEEEoverridecommandlockouts 
\begin{document}
\title{\LARGE \bf 
Data-Driven Physics Embedded Dynamics with Predictive Control and Reinforcement Learning for Quadrupeds
%
}
\author{
Prakrut Kotecha$^*$$^{1}$, Aditya Shirwatkar$^*$$^{1}$, Shishir Kolathaya$^{2}$
\thanks{$^*$ Equal Contribution, This project is funded by ARTPARK and Kotak IISc AI-ML Centre (KIAC)-Google}
\thanks{$^{1}$A. Shirwatkar, P. Kotecha, are with the Robert Bosch Center for Cyber-Physical Systems, Indian Institute of Science, Bengaluru.}%
\thanks{$^{2}$S. Kolathaya is with the Robert Bosch Center for Cyber-Physical Systems and the Department of Computer Science \& Automation, Indian Institute of Science, Bengaluru.}
\thanks{Email: \href{mailto:stochlab@iisc.ac.in}{stochlab@iisc.ac.in}, Website: \href{https://www.stochlab.com/PEPC/}{link}}%
}

\maketitle
\thispagestyle{empty}
\pagestyle{empty}

\begin{abstract}

State-of-the-art quadrupedal locomotion approaches integrate Model Predictive Control (MPC) with Reinforcement Learning (RL), enabling complex motion capabilities with planning and terrain-adaptive behaviors. However, they often face compounding errors over long horizons and have limited interpretability due to the absence of physical inductive biases. We address these issues by integrating Lagrangian Neural Networks (LNNs) into an RL–MPC framework, enabling physically consistent dynamics learning. At deployment, our inverse-dynamics infinite-horizon MPC scheme avoids costly matrix inversions, improving computational efficiency by up to 4× with minimal loss of task performance. We validate our framework through multiple ablations of the proposed LNN and its variants. We show improved sample efficiency, reduced long-horizon error, and faster real-time planning compared to unstructured neural dynamics. Lastly, we also test our framework on the Unitree Go1 robot to show real-world viability.

\end{abstract}



\section{Introduction}

The ability to accurately model and predict system dynamics is important for planning and control in quadruped robots. Traditional approaches derive explicit equations of motion using first principles, yielding interpretable and physically consistent models \cite{idto, jumpmpc, ci_mpc}. However, they can be computationally intensive and require precise knowledge of system parameters, which may not always be available or practical to obtain. In complex systems, such as legged robots, the interplay of multiple degrees-of-freedom (DoF), nonlinear interactions, and external disturbances further complicates the task of deriving accurate dynamics models.

Advances in machine learning have introduced data-driven methods as an alternative to classical dynamics modeling. We refer to them as Opaque Neural Network (ONN) dynamics in our context. These approaches leverage large datasets to learn system behavior without explicit reliance on laws of physics \cite{chen2018neural, gpr1}. However, these models lack interpretability, struggle to generalize to unseen scenarios, and accumulate long-horizon errors that degrade control performance. Here, lack of interpretability means that the learned dynamics cannot explicitly map to physical quantities like generalized coordinates, energies, and mass–inertia matrices.

\begin{figure}[htp!]
    \captionsetup{font=footnotesize}
    \centering
    \includegraphics[width=0.8\linewidth]{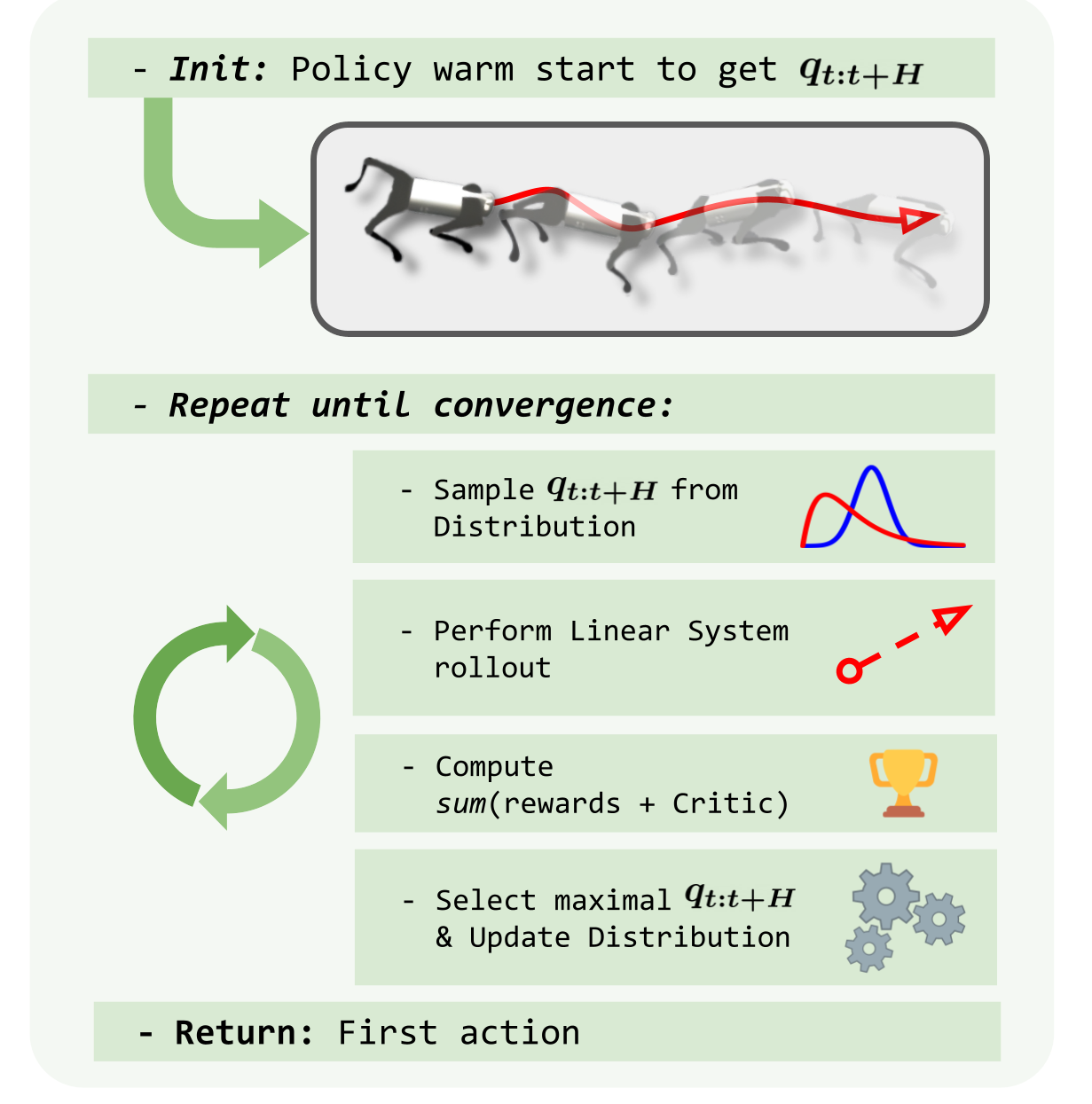}
    \caption{Overview of our inverse dynamics-based planner. At each control step, the method samples candidate joint trajectories, computes the corresponding torques, evaluates them using learned reward and value functions, and updates the distribution for the next sampling round.}
    \label{fig:overview}
    \vspace{-18pt}
\end{figure}

This limitation has spurred interest in hybrid approaches that combine the strengths of physics-based modeling and data-driven learning \cite{raissi2019physics, djeumou2022neural}.
Lagrangian Neural Networks (LNNs) \cite{LNN, cartesianLNN} represent one such direction, where instead of fitting dynamics directly, the model parameterizes the system’s Lagrangian and derives equations of motion via the Euler-Lagrange formulation.
LNNs embed Lagrangian mechanics into learned dynamics, producing physically consistent, interpretable predictions that improve generalization and reliability for planning.
A related paradigm, Hamiltonian Neural Networks (HNNs) \cite{HNN}, similarly enforces physical structure, and can be reformulated as an LNN through a Legendre transformation. In this work, we focus on LNNs due to their suitability for the control problem at hand, while noting the close theoretical connection between the two approaches.

\begin{table*}[htp!]
    \vspace{3pt}
    \scriptsize
    \centering
    \captionsetup{font=footnotesize}
    \caption{Comparison of proprioceptive quadruped locomotion methods, including our approach. The table summarizes key trade-offs across inference speed (higher rank is faster), planning capability, model interpretability, and terrain generalization.}

    \vspace{-3pt}
    \label{tab:comparison}
    \setlength{\arrayrulewidth}{1.5pt} 
    \renewcommand{\arraystretch}{1.3} 
    \begin{tabular}{c | c  c  c  c}
        \hline
        \thead{\textbf{Quadruped Locomotion Methods}} & \thead{\textbf{Inference Speed (1-5)}} & \thead{\textbf{Planning}} & \thead{\textbf{Interpretable Dynamics}} & \textbf{Multi-Terrain}\\ 
        \hline
        Nonlinear MPC \cite{idto, jumpmpc, ci_mpc} & 3 to 4 & \green{\cmark} & \green{\cmark} & \red{\xmark}   \\
        DreamWaQ \cite{dreamwaq}, HIMLoco \cite{himloco} & 5      & \red{\xmark}   &  None          & \green{\cmark} \\
        PIP-Loco \cite{piploco}                          & 4      & \green{\cmark} & \red{\xmark}   & \green{\cmark} \\
        Ours                                             & 3      & \green{\cmark} & \green{\cmark} & \green{\cmark} \\
        \hline
    \end{tabular}
    \vspace{-10pt}
\end{table*}

The application of physics-informed learning frameworks to real-world systems, particularly in robotics, remains an active area of research. Deep Lagrangian Networks (DeLaN) \cite{delan, delan2} was among the first works to demonstrate the viability of LNNs in low DoF systems using model-based control. 
LNNs have also been applied in model-based RL \cite{lmbrl2}, however, the computational cost of real-time inference limits their use in high-DoF systems like quadrupeds, where contact dynamics and underactuation.

To address these limitations, there exist hybrid approaches that combine the structure of model-based control with the flexibility of model-free RL. In particular, integrating RL policies with Model Predictive Control (MPC) has significantly improved the performance of complex dynamical systems \cite{piploco, loop, tdmpc2}. 
Here for training, they commonly employ a Dreamer module \cite{piploco} to jointly learn an ONN dynamics model, a reward estimator, a policy and a value function for long-horizon control. For clarity, please note
that the Dreamer module \cite{piploco} is different from the Dreamer algorithm \cite{dreamer}. The Dreamer module is used to facilitate future state prediction and planning during deployment (see Section \ref{sc:training_dreamer} for more details). During inference, an approximation of the infinite-horizon trajectory optimization problem via bootstrapping of the value function, is solved by warm-starting it using the RL policy for faster convergence. 
While effective, the ONN-based dynamics suffer from generalization and long-horizon errors, motivating our use of an LNN-based counterpart. Further, since LNNs face practical challenges for real-time inference, their direct integration cannot be done in RL-MPCsetups. Hence, there is a pressing need for methods that strike a balance between computational efficiency and physical consistency, enabling robust real-time planning and control in such demanding systems. 

To address the challenges mentioned above, we propose a framework that integrates LNNs with RL-MPC framework for real-time quadrupedal locomotion. Specifically, the transition dynamics are learned concurrently as the model-free RL policy explores and are subsequently employed for planning during deployment. Table \ref{tab:comparison} summarizes how our approach compares with other propriocepive quadruped locomotion methods in terms of efficiency, planning capability, interpretability, and terrain adaptability.  An overview of our inverse dynamics-based planner is shown in Fig. \ref{fig:overview}.
Overall, our approach enhances the sample efficiency of the dynamics model while reducing rollout errors during deployment.
To summarize, our key contributions include:


\begin{enumerate}
\item \textbf{LNN-based RL-MPC Architecture} - We present the first framework integrating physics-consistent LNNs into an RL-MPC pipeline for locomotion, enabling physically interpretable long-horizon predictions.
\item \textbf{Planning with Learned Inverse Dynamics} - We formulate an inverse dynamics optimization over joint trajectories that eliminates mass matrix inversions at deployment, achieving up to $4\times$ lower latency than forward-dynamics LNN planners while maintaining competitive returns enabling real-time deployment.
\item \textbf{Empirical validation in sim and real world} - We benchmark across 5 dynamics model variants and validate on a Unitree Go1 across 6 terrain types.
\end{enumerate}


\section{Setup}  
\label{sc:setup}

In this section, we present the mathematical setup, starting with the underlying problem formulation, followed by an overview of the RL training pipeline.

\subsection{Preliminaries}
\label{sc:preliminaries}

We formulate our problem as an infinite-horizon discounted Partially Observable Markov Decision Process (POMDP), defined by the tuple: $\mathcal{M} = \{\mathcal{S}, \mathcal{A}, \mathcal{O}, r, \mathcal{P}, \gamma, \mathcal{T}\}$,
where $\mathcal{S} \subset \mathbb{R}^n$ denotes the state space, $\mathcal{O} \subset \mathbb{R}^p$ the observation space, and $\mathcal{A} \subset \mathbb{R}^m$ the action space. The observation history over the past $M$ steps is denoted as $\mathcal{O}^M \subset \mathbb{R}^{p \times M}$. The system evolves according to the transition function $\mathcal{P}: \mathcal{S} \times \mathcal{A} \mapsto \Pr(\cdot)$, where $\Pr$ denotes a probability distribution over successor states. Partial observability is governed by the observation function $\mathcal{T}: \mathcal{S} \mapsto \mathcal{O}$. The agent's behavior is guided by a reward function $r : \mathcal{S} \times \mathcal{A} \mapsto \mathbb{R}$, with $\gamma \in (0,1)$ denoting the discount factor. Further, we represent the full state of the quadruped using $[q, \dot q]$, where $q \in \mathbb{R}^{18}$ denotes the generalized positions and $\dot q \in \mathbb{R}^{18}$ the generalized velocities. Input actuation in the form of torques is denoted by $u \in \mathbb{R}^{12}$.

\subsubsection{Observations \& Actions}

The observation vector $o \in \mathcal{O}$ comprises a combination of proprioceptive signals. Specifically, it includes joint angle $\Theta \in \mathbb{R}^{12}$, joint velocities $\dot{\Theta} \in \mathbb{R}^{12}$, projected gravity $\text{g}_p$, and the previous action $a_{t-1}$. Additionally, desired linear and angular velocity commands for the base, $\text{v}^{cmd}$, are provided to the policy.

The action vector $a \in \mathcal{A}$ represents joint angle offsets applied to the nominal joint angles. The commanded joint positions at time step $t$ are computed as $\Theta^{\text{cmd}}_t = a_t + \Theta_{\text{nominal}}$. These commanded positions are passed through an actuator network \cite{eth2019} to generate the required joint torques ($u$). 


\subsubsection{Reward Function}

The reward function is designed to promote desirable locomotion behaviors aligned with the control objectives. These objectives include accurately tracking the commanded linear and angular velocities, maintaining a desired body height, and preserving a stable body orientation. A comprehensive description of the specific reward terms can be found in \cite{leggedgym}.

\begin{figure}[htp!]
    \captionsetup{font=footnotesize}
    \centering
    \includegraphics[width=0.9\linewidth]{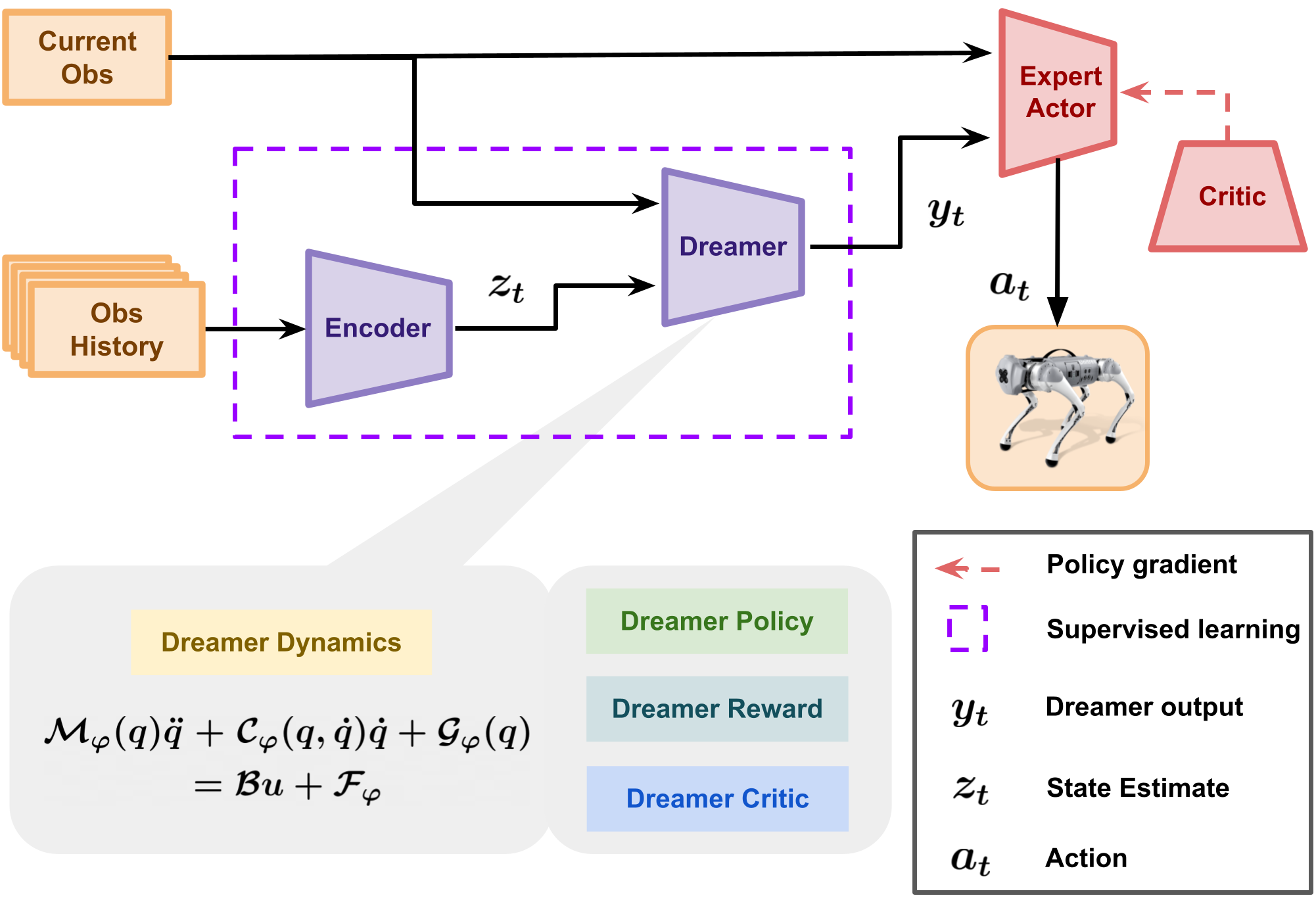}
    \caption{Our training framework where observation history is encoded into full-state estimates, and passed through a physics-informed Dreamer module to augment expert actor with future predictions for robust policy learning.}
    \label{fig:RL_flowchart}
    \vspace{-10pt}
\end{figure}

\subsection{RL Training} 
\label{sc:training}

Our training design is based on an Asymmetric Actor-Critic \cite{asymmetric} formulation of model-free RL from \cite{piploco}. Here, an expert actor receives augmented inputs from a Dreamer module, and the critic takes privileged observations. The privileged observation comprises the height field around the robot, external forces, the robot's linear velocity, and other unobservable data, in addition to the proprioceptive observations.
The motivation for this is to address the information asymmetry between the actor and the critic. 

However, to our end, we make two key changes: (i) the encoder is trained to output full-body state estimates, and (ii) the Dreamer module is physics-informed through the use of LNNs (see Section \ref{sc:methods} for more details).
We provide an overview of this approach in Fig. \ref{fig:RL_flowchart}.
Running MPC planners during every training rollout across thousands of parallel environments \cite{isaac_gym, leggedgym} is both slow and memory-intensive, making recent model-based RL methods \cite{loop, tdmpc2} impractical in our setting.
To avoid this, we keep it simple and efficient by deferring planning to the deployment phase, similar to \cite{piploco}, where its benefits can be fully leveraged without affecting training time.

\section{Methodology}
\label{sc:methods}
The previous section established the training pipeline. We now detail its three technical components: the LNN dynamics model, the encoder and Dreamer module built on top of it, and the deployment-time inverse-dynamics planner.

\subsection{Lagrangian Neural Networks (LNNs)}

Consider the Lagrangian, defined as the difference between the kinetic and potential energies: 
\(\mathcal{L} = \frac{1}{2} \dot q^T \mathcal{M}(q) \dot q - \mathcal{V}(q)\).
The system's equations of motion follow from the Euler-Lagrange equation:

{\small \vspace{-2pt}
\begin{equation*}
\mathcal{M}(q) \ddot{q} + \mathcal{C}(q, \dot{q}) \dot{q} + \mathcal{G}(q) = \mathcal{B} u + \mathcal{H}(q,\dot q),
\end{equation*}}
where \(\mathcal{M}\) is the mass matrix, \(\mathcal{C}\) is for Coriolis and centrifugal forces, \(\mathcal{G}\) represents gravitational forces, \(\mathcal{H}\) denotes external forces, and \(\mathcal{B}\) accounts for actuation mapping at the DoFs.
\(\mathcal{C}(q, \dot{q})\) and \(\mathcal{G}(q)\) are derived from the partial derivatives, as follows:

{\small \vspace{-12pt}
\begin{equation*}
\mathcal{C}(q, \dot q) = \nabla_q\mathcal{M}(q) \ \dot q - \frac{1}{2}\nabla_q(\dot q^T \mathcal{M}(q)),\quad \mathcal{G}(q) = \nabla_q\mathcal{V}(q).
\end{equation*}}

To embed this physical structure into a learnable dynamics model, we adopt the DeLaN framework, which models dynamics directly via the Lagrangian formulation. We parametrize \(\mathcal{M}(q)\), \(\mathcal{V}(q)\), and \(\mathcal{H}(q, \dot q)\) using neural networks with parameters \(\varphi\). 
However, this requires careful handling to ensure physical consistency, as discussed below.

\subsubsection{Mass Matrix Parameterization}
Following \cite{delan, delan2}, the mass matrix is parameterized as a symmetric positive-definite (SPD) matrix: $\mathcal{M}_\varphi(q) = \mathcal{Y}_\varphi(q)\mathcal{Y}_\varphi(q)^T + \varepsilon I$, where \( \mathcal{Y}_\varphi(q) \) is a learnable lower triangular matrix with entries produced by a multi-layer perceptron (MLP), and \(\varepsilon I\) is a small diagonal matrix added for numerical stability.

\subsubsection{Next State Prediction}\label{sc:next_state_pred}
The system's continuous-time evolution can be expressed in state-space form:

{\small \vspace{-12pt}
\begin{equation*}
\frac{d}{dt}
\begin{bmatrix}
q \\
\dot q
\end{bmatrix}
=
\begin{bmatrix}
\dot{q} \\
\mathcal{M}_\varphi^{-1}(q) \left( \mathcal{B} u + \mathcal{H}_\varphi(q, \dot q) - \mathcal{C}_\varphi(q, \dot{q}) \dot{q} - \mathcal{G}_\varphi(q) \right)
\end{bmatrix}.
\end{equation*}}
For practical implementation, the dynamics are discretized using first-order integration. Given the state vector \([q_t, \dot q_t]\) and acceleration \(\ddot{q}_t\) at time \(t\), the update to \(t+1\) is:

{\small \vspace{-2pt}
\begin{equation}
\begin{bmatrix}
q_{t+1} \\
\dot q_{t+1}
\end{bmatrix}
=
\begin{bmatrix}\label{eq:Forward}
q_t + \dot q_t \Delta t + \ddot q_t \Delta t^2 \\
\dot q_t + \ddot q_t \Delta t
\end{bmatrix}
\equiv f_\varphi(q_t, \dot q_t, u_t).
\end{equation}}
Here, \(\Delta t\) denotes the timestep size, and \(f_\varphi\) represents the forward dynamics map. 
The control input \(u\) is generated via an actuator network \cite{eth2019}. This formulation ensures physically consistent state evolution while remaining computationally efficient. 
The forward dynamics map \(f_\varphi\) defined above requires the full physical state $[q, \dot q]$ as input. Since this state is not directly observable on hardware, we next describe the encoder that estimates it from proprioceptive history, and the Dreamer module that uses \(f_\varphi\) to generate physics-informed imagined trajectories for policy training.

\subsection{Encoder \& Dreamer Module}
\label{sc:dreamer_module}

To enable the expert actor to reason about the future, we augment its input using the predicted next full state. This is done through the Dreamer module, which consists of a learned dynamics model \( d_\varphi \), policy \( \pi_\varphi \), reward function \( r_\varphi \), and value function \( V_\varphi \). All modules operate on full state information \( z_t = (q_t, \dot{q}_t) \in \mathbb{R}^{36} \), which is not directly observed and must be inferred. This setup is conceptually related to \cite{i2a} and \cite{piploco}, but here the dreamt trajectory is generated from a physics-informed model (LNN), ensuring consistent behavior even under out-of-distribution transitions.

\subsubsection{Full-State Estimation}
We learn an encoder network \( \mathcal{E}_\varphi: \mathcal{O}^M \rightarrow \mathbb{R}^{36} \) that maps a history of proprioceptive observations \( o_{t-M:t} \) to the full physical state \( z_t \). The encoder is trained via supervised regression using simulation ground-truth states: $\mathcal{L}_{\text{Encoder}} = \left\| \mathcal{E}_\varphi(o_{t-M:t}) - z_t \right\|^2$

\begin{algorithm}[htp!]
\captionsetup{font=footnotesize}
\caption{Physics-Informed Dreaming}
\label{alg:dreamer_rollout}
\footnotesize
\begin{algorithmic}[1]
    \Require Observation history \( o_{t-M:t} \), horizon \( H \)
    \State \( z_t \gets \mathcal{E}_\varphi(o_{t-M:t}) \)
    \State \( \Tilde{z}_t \gets z_t \), \quad \( y_t \gets \{ \Tilde{z}_t \} \)
    \For{ \( k = 0 \) to \( H-1 \) }
        \State \( a_k \gets \pi_\varphi(\Tilde{z}_{t+k}) \)
        \State \( \tilde{z}_{t+k+1} \gets d_\varphi(\tilde{z}_{t+k}, a_k) \)
        \State \( y_t \gets y_t \cup \{ \Tilde{z}_{t+k+1} \} \)
    \EndFor
    \State \Return Dreamt trajectory \( y_t = \{ \tilde{z}_{t+1}, \dots, \tilde{z}_{t+H} \} \)
\end{algorithmic}
\end{algorithm}

\subsubsection{Physics-Informed Dreaming}
The estimated state \( z_t \) is used to generate a predicted trajectory \( y_t \), where each future state is obtained by rolling out the Lagrangian dynamics model \( d_\varphi \) under the dreamer policy \( \pi_\varphi \) (see Algorithm~\ref{alg:dreamer_rollout}). Here, \( d_\varphi(z, a) \) coincides with \( f_\varphi \) in Equation \eqref{eq:Forward} when applied together with an actuator network \cite{eth2019}, which converts actions \( a \) into the corresponding control inputs \( u \) (torques) required by the dynamics. The resulting dreamt trajectory is then combined with the current proprioceptive observations and provided as an augmented input to the expert actor.

\subsubsection{Supervised Training of Dreamer Module}
\label{sc:training_dreamer}
\label{sc:dreamer_loss}

All networks within the Dreamer module are trained using supervised targets. For each timestep \( t \) in the replay buffer, we construct the training tuple \( (o_{t-M:t}, a_t^{\text{expert}}, r_t^{\text{sim}}, z_{t+1}^{\text{sim}}, V_{t}^{\text{target}}) \), where: \( a_t^{\text{expert}} \) is the action from the expert actor, \( r_t^{\text{sim}} \) and \( z_{t+1}^{\text{sim}} \) are the ground-truth reward and next state from simulation, \( V_t^{\text{target}} \) is a value target provided by the privileged critic. Then each component is trained to minimize the following loss:

{\small \vspace{-10pt}
\begin{align*}
\mathcal{L}_{\text{Dreamer}} = 
&\underbrace{
    \left\| d_\varphi(z_t, a_t^{\text{expert}}) - z_{t+1}^{\text{sim}} \right\|^2
}_{\text{\footnotesize Dreamer Dynamics}} 
+
\underbrace{
    \left\| r_\varphi(z_t, a_t^{\text{expert}}) - r_t^{\text{sim}} \right\|^2
}_{\text{\footnotesize Dreamer Reward}} \notag \\
&+
\underbrace{
    \left\| V_\varphi(z_t) - V_t^{\text{target}} \right\|^2
}_{\text{\footnotesize Dreamer Critic}} 
+
\underbrace{
    \left\| \pi_\varphi(z_t) - a_t^{\text{expert}} \right\|^2
}_{\text{\footnotesize Dreamer Policy}}
\label{eq:dreamer_loss_fixed}
\end{align*}}
This loss is computed independently for each sample \( t \) in the offline buffer, and gradients are propagated only through the respective module. The use of supervised learning ensures sample-efficient training without backpropagation through imagined rollouts or the Proximal Policy Optimization (PPO) \cite{ppo} loss of the expert actor. 

It is important to note the distinct roles of the two learned policies: the expert actor, trained via PPO, is the sole policy that interacts with the environment during training. 
The Dreamer policy warm-starts the inverse-dynamics MPC solver at inference time alongside the previous MPC solution and is never executed as a standalone controller. Further, the expert actor's role is confined entirely to training.
From now on, for simplicity, we omit the timestep subscripts in the following section on deployment.

\subsection{Deployment: Planning with Inverse Dynamics}
\label{sc:algo}

With the encoder and Dreamer module trained, we adopt a planning approach during inference. The optimization objective, using full state \( z = [q, \dot{q}] \), is expressed as:

{\small \vspace{-10pt}
\begin{equation}
\label{eq:Forward_MPC}
\begin{aligned}
    \max_{a_{0:H-1}} \quad & \sum_{k=0}^{H-1} \gamma^k r_\varphi(z_k, a_k) + \gamma^H V_\varphi(z_H) \\
    \text{s.t.} \quad & z_0 = z_{\text{init}}, \\
    & z_{k+1} = d_\varphi(z_k, a_k), \quad \forall k = 0, \dots, H-1.
\end{aligned}
\end{equation}}

Solving this optimization yields an \(H\)-step action sequence, using the learned reward and value functions from Section \ref{sc:dreamer_module}. The sequence begins from the initial state \( z_{\text{init}} \) and evolves according to the learned dynamics model \( d_\varphi \).

\subsubsection{Computational Challenges: Forward vs. Inverse Dynamics}

Although forward dynamics MPC policies have been successfully demonstrated on manipulators using DeLaN \cite{context_delan}, applying them directly to Equation \eqref{eq:Forward_MPC} introduces substantial computational challenges. These arise from repeated mass matrix inversions during the forward pass. Even when exploiting the SPD structure, matrix decompositions remain a significant bottleneck, especially for sampling-based optimization algorithms that evaluate many trajectories in parallel. Consequently, direct forward-dynamics-based integration becomes impractical for high-dimensional systems, such as quadrupeds, in real-time settings.

\begin{algorithm}[htp!]
\footnotesize
\captionsetup{font=footnotesize}
\caption{Inverse Dynamics-Based $\infty$ Horizon MPC}
\label{alg:Inverse_MPC}
\begin{algorithmic}[1]
\Require 
$(\mu^{\mathrm{prev}}, \sigma^{\mathrm{prev}})$: Previous MPC solution; $z = [q, \dot{q}]$: Full state estimate from encoder;
\State Generate $(\mu^{\mathrm{RL}}, \sigma^{\mathrm{RL}})$ from the Dreamer rollout starting at $z$
\State $(\mu_0, \sigma_0) \gets \alpha (\mu^{\mathrm{prev}}, \sigma^{\mathrm{prev}}) + (1 - \alpha)(\mu^{\mathrm{RL}}, \sigma^{\mathrm{RL}})$

\For{$i = 1$ to $N$}
    \State $A \gets \emptyset$, $R \gets \emptyset$

    \State \textit{\grey{// Sample trajectories}}
    \vspace{-5pt}\[
    q^{(j)}_{1:H} \sim \mathcal{N}(\mu_{i-1}, \sigma_{i-1}), \quad \forall j \in [1, M]
    \]
    \vspace{-10pt}\[
    q^{(j)}_{1:H} \sim \mathcal{N}(\mu^{\mathrm{RL}}, \sigma^{\mathrm{RL}}), \quad \forall j \in [M+1, M + M_\pi]
    \]
    \vspace{-12pt}

    \ForAll{$j$ in sampled trajectories}

        \State $\bar{R}^{(j)} = 0$
        \For{$k = 1$ to $H - 1$}
            \State $(\dot{q}^{(j)}_k , \ddot{q}^{(j)}_k) = \mathcal{F}(q^{(j)}_{k-1}, q^{(j)}_k, q^{(j)}_{k+1})$

            \State $u^{(j)}_k = g_\varphi(q^{(j)}_k, \dot{q}^{(j)}_k, \ddot{q}^{(j)}_k)$
            
            \State $\bar{R}^{(j)} = \bar{R}^{(j)} + \gamma^k r_\varphi(\cdots) - \lambda \cdot \text{RootForcePenalty}(\mathcal{B}u^{(j)}_k)$
        \EndFor
        
        \State $\bar{R}^{(j)} = \bar{R}^{(j)} + \gamma^H V_\varphi(q^{(j)}_H, \dot{q}^{(j)}_H)$
                
        \State $R \gets R \cup \{\bar{R}^{(j)}\}$, \quad $A \gets A \cup \{q^{(j)}_{1:H}\}$
    \EndFor

    \State $A_E \gets$ Top $M_{\text{elite}}$ trajectories from $A$ by return $R$
    \State Fit Gaussian $(\mu_\text{elite}, \sigma_\text{elite})$ to $A_E$
    \State $(\mu_i, \sigma_i) \gets \beta (\mu_\text{elite}, \sigma_\text{elite}) + (1 - \beta)(\mu_{i-1}, \sigma_{i-1})$
\EndFor

\State \Return First action from $\mathcal{N}(\mu^N, \sigma^N)$
\end{algorithmic}
\end{algorithm}

To address this, we propose an alternative optimization formulation expressed directly in terms of joint positions \(q\), inspired by \cite{idto, goal_dd}:

{\small \vspace{-5pt}
\begin{equation} \label{eq:Inverse_MPC}
\begin{aligned}
\max_{q_{1:H}} \quad & \sum_{k=0}^{H-1} \gamma^k r_\varphi(q_k, \dot{q}_k, u_k) 
+ \gamma^H V_\varphi(q_H, \dot{q}_H) \\
\text{s.t.} \quad & q_0 = q_{\text{init}}, \quad \dot{q}_0 = \dot{q}_{\text{init}}, \\
& (\dot{q}_k, \ddot{q}_k) = \mathcal{F}(q_{k-1}, q_k, q_{k+1}), \\
& u_k = g_\varphi(q_k, \dot{q}_k, \ddot{q}_k), \quad \forall k = 1,\dots,H-1.
\end{aligned}
\end{equation}}

Here, the reward function \( r_\varphi \) is defined directly in terms of \( u \) instead of the policy action \( a \), allowing inverse dynamics to be explicitly incorporated. The inverse dynamics model \( g_\varphi \) is defined as:

{\small \vspace{-10pt}
\begin{equation*} \label{eq:inverse}
    \mathcal{B} u = \mathcal{M}_\varphi(q) \ddot{q} + \mathcal{C}_\varphi(q, \dot{q}) \dot{q} + \mathcal{G}_\varphi(q) - \mathcal{H}_\varphi(q, \dot q) \equiv g_\varphi(q, \dot q, \ddot q).
\end{equation*}}
Finally, the finite-difference operator \( \mathcal{F} \) provides approximations of velocity and acceleration:

{\small \vspace{-10pt}
\begin{equation*} \label{eq:F_operator}
\mathcal{F}(q_{k-1}, q_k, q_{k+1}) 
= \left( 
\frac{q_k - q_{k-1}}{\Delta t}, \quad 
\frac{q_{k+1} - 2q_k + q_{k-1}}{\Delta t^2} 
\right).
\end{equation*}}

See Algorithm \ref{alg:Inverse_MPC} for a detailed description of solving Equation \eqref{eq:Inverse_MPC}.






\subsubsection{Pros and Cons of Inverse Dynamics}
An advantage of this formulation is that future states evolve through a simple linear rollout, with all nonlinearities encapsulated in the reward calculation via \(u\). Equation \eqref{eq:Forward_MPC} requires optimizing over \(a\) and rely on forward dynamics for state propagation, which introduces the aforementioned computational burdens. By leveraging inverse dynamics, we optimize directly over the joint trajectory \(q\), bypassing explicit matrix inversions.

However, for underactuated systems, it is not immediately obvious that the so-called \emph{root forces} components of \(\mathcal{B}u\) should be zero in the optimal solution, since unactuated DoFs cannot generate torque or forces to evolve the state. Ensuring this property is particularly challenging when using the sampling-based optimization method we adopt. Gradient- or Hessian-based constrained optimization solvers can enforce this condition \cite{idto}, but they introduce additional computational overhead due to the need for gradient calculations of the learned function approximators. Instead, in our approach, we simply discard trajectories during the sampling step if they exhibit high root forces. 
However, we note that the frequency of trajectory rejection due to root force violations, and sensitivity to the penalty weight $\lambda$, remain open questions that warrant systematic study in future work.

\section{Results}
\label{sc:results}

In this section, we present the implementation details, followed by a comprehensive evaluation of our framework against baseline methods. We report results on training performance to capture sample efficiency, and deployment-time effectiveness to assess prediction accuracy, inference speed and performance. We validate our controller on real hardware, for robustness and generalization with minimal simulation-to-reality transfer issues. 

\begin{table}[htp!]
    \footnotesize
    \centering
    \captionsetup{font=footnotesize}
    \caption{Hyperparameters used for training and deployment. Separate configurations are used for the RL and MPC modules to balance performance and efficiency.}
    \label{tab:mpc}
    \vspace{-3pt}
    \setlength{\arrayrulewidth}{1.5pt}
    \begin{tabular}{c c}
        \hline
        \textbf{Hyperparameters} & \textbf{Value} \\
        \hline
        \\[-4pt]
        \multicolumn{2}{c}{\textbf{\textit{Training}}} \\[1pt]
        Hidden layers for RL & $\begin{bmatrix} 512, 256, 128 \end{bmatrix}$ \\
        Hidden layers for Dreamer & $\begin{bmatrix} 256, 256 \end{bmatrix}$ \\
        Activation function & \texttt{tanh} \\
        Number of environments & 4096 \\
        Number of mini batches & 25 \\
        Number of steps per environment & 100 \\
        \hline
        \\[-4pt]
        \multicolumn{2}{c}{\textbf{\textit{Deployment}}} \\[1pt]
        Planning Horizon ($H$) & 8 \\
        Number of MPPI Iterations ($N$) & 6 \\
        Number of Sampled Trajectories ($M$) & 500 \\
        Number of RL Policy Samples ($M_\pi$) & 30 \\
        Number of Elite Trajectories ($M_\text{elite}$) & 60 \\
        Discount Factor ($\gamma$) & 0.99 \\
        Temporal Momentum ($\beta$) & 0.95 \\
        Initialization Mix Factor ($\alpha$) & 0.5 \\
        Constraint Penalty Weight ($\lambda$) & 1.0 \\
        \hline
    \end{tabular}
\end{table}

\subsubsection{Implementation Details}
In the RL framework, we employ an asymmetric actor-critic architecture implemented as an MLP. Similarly, the encoder and Dreamer module have been implemented as an MLP; the hyperparameter details of the same are described in Table \ref{tab:mpc}. During training, privileged information about the next state is stored in the buffer, enabling the training of the dynamics model. The base RL algorithm is PPO \cite{ppo} with modifications tailored for our quadruped environment \cite{leggedgym}. For deployment, we integrate the Dreamer module into our sampling-based MPC framework (section \ref{sc:algo}), whose hyperparameters are also listed in Table \ref{tab:mpc}.

\subsubsection{System Details} For training, we utilized an open-source massively parallel RL environment setup for the Unitree Go1 quadruped robot, based on Nvidia's Isaac Gym simulator \cite{isaac_gym, leggedgym}. The neural networks for PPO were implemented using PyTorch \cite{pytorch}, while the Dreamer module was implemented using JAX \cite{jax}. Training was conducted on a desktop system with an Intel Xeon(R) Gold 5318Y CPU (48 cores, 2.10 GHz), 512 GB of RAM, and an NVIDIA RTX 6000 Ada Generation GPU. For hardware deployment, we used a system with an AMD Ryzen 9 7940HS @ 3.992 GHz, 32 GB of RAM, and an Nvidia GeForce RTX 4060 Laptop GPU.

\subsubsection{Baselines}

We compare our method against a range of baselines, including models without any inductive biases, LNN \cite{LNN, delan}, and its variants introduced for quadrupedal planning \cite{LNN_AIR}. These baselines are categorized based on the type of dynamics modeling (forward or inverse), the state space structure incorporated, and the strategy used during training and inference.

\textbf{ONN (Opaque Neural Network)} serves as our unstructured baseline. It directly maps current observations and actions to the next observation using a feedforward neural network, without incorporating any physics priors:

{\small \vspace{-2pt}
\[ 
\hat{o}_{t+1} = d^\text{ONN}_\varphi(o_t, a_t), \quad \mathcal{L}_\text{ONN} = \left\| o_{t+1} - \hat{o}_{t+1} \right\|^2.
\]}
This baseline allows us to assess the effect of imposing physical structure on dynamics learning. We exclude baselines such as LSTMs, as our framework operates on an estimated full state from a fixed observation window, making recurrence unnecessary. Moreover, these tend to obscure internal representations, making them less suitable for our goal of interpretable dynamics modeling.

\textbf{LNN Forward} \cite{LNN} follows the standard Lagrangian Neural Network formulation, where the mass matrix is fully learned and inverted during both training and inference. This baseline evaluates the benefits and trade-offs of learning structured dynamics without the diagonalization simplification.


\textbf{DeLaN} \cite{delan} also models system dynamics using the Lagrangian formulation. However, unlike LNN Forward, it diagonalizes the mass matrix during both training and inference, significantly reducing computational complexity while maintaining physical consistency.


\textbf{LNN Inverse} uses an inverse dynamics formulation during training to predict joint torques required to produce observed state transitions. 
This design highlights how structured training can influence generalization, even when inference follows a different formulation:

{\small \vspace{-10pt}
\[
\mathcal{L}_\text{LNN-Inverse} =
\left\| \tau_t - \hat{\tau}_t \right\|^2
+ \left\| \text{RootForce}(\mathcal{B}\hat{\tau}_t) \right\|^2,
\quad \mathcal{B} \hat{\tau}_t = g_\varphi(\cdots)
\]}
\textbf{CoM LNN} simplifies the system representation by modeling the robot as a single rigid body and applying Lagrangian dynamics to predict the center of mass (CoM) evolution. It is trained using the following loss: $\mathcal{L}_\text{CoM-LNN} = \left\| c_{t+1} - \hat{c}_{t+1} \right\|^2$, where $c_{t+1}$ is the ground-truth CoM state, and $\hat{c}_{t+1}$ is the predicted value obtained from CoM-based dynamics. This baseline isolates the effect of coarse-grained modeling and is useful for understanding the role of high level abstraction in learned dynamics.

\begin{figure}[htp!]
    \centering
    \captionsetup{font=footnotesize}
    \includegraphics[width=0.8\linewidth]{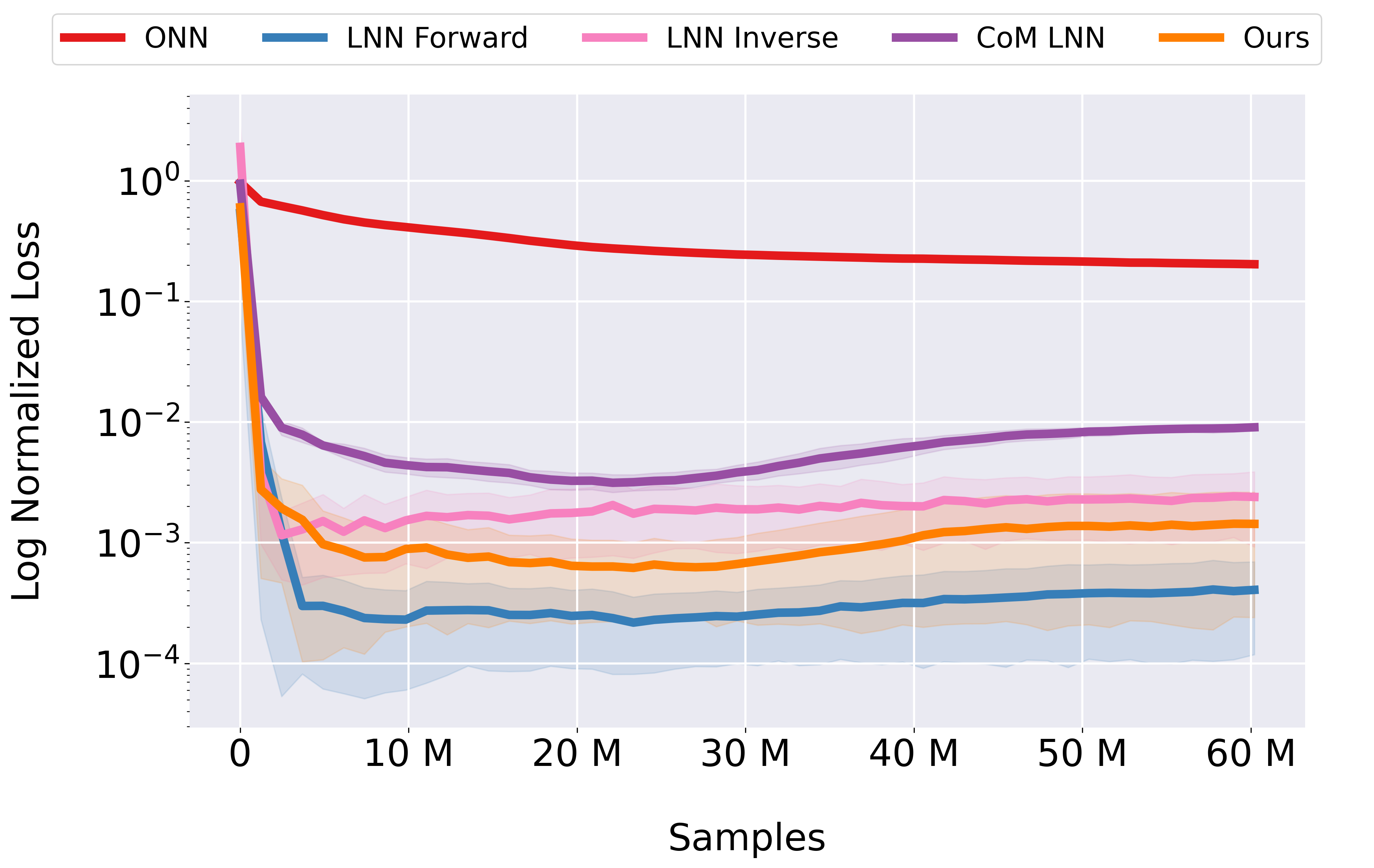}
    \caption{Training loss curves across different dynamics models. Our method achieves strong sample efficiency while balancing compute cost, compared to LNN Forward and ONN baselines.}
    \label{fig:dynamics_loss}
    \vspace{-10pt}
\end{figure}

\subsection{Training and Sample Efficiency}

We evaluate model learning efficiency from two perspectives: \textbf{training efficiency}, measured as the computational cost per iteration, and \textbf{sample efficiency}, defined as the number of samples required to reach convergence. 
Fig. \ref{fig:dynamics_loss} shows the normalized training loss over 60 million environment steps. While ONN achieves lower per-iteration compute cost ($\approx$ 4.5 secs per iteration), its loss remains significantly higher throughout training, indicating poor sample efficiency. In contrast, all the LNN-based methods converge faster. LNN Forward (with full mass matrix inversion) achieves the lowest training loss overall but at a high computational cost ($\approx$ 22 secs per iteration), making it less suited for real-time applications.

Our method (which is equivalent to DeLaN in training) exhibits lower training loss because of physics priors, while also keeping the per-iteration cost low using diagonalization while training ($\approx$ 17.5 secs per iteration). Our approach achieves comparable loss to LNN Inverse while also offering advantages in inference speed and planning robustness.

These results confirm that incorporating physical priors and Lagrangian structure improves sample efficiency. While structured models introduce additional overhead, this cost is compensated by faster convergence and improved long-term accuracy, particularly in high-dimensional tasks such as quadrupedal locomotion.





\begin{figure}[htp!]
    \centering
    \captionsetup{font=footnotesize}
    \includegraphics[width=0.8\linewidth]{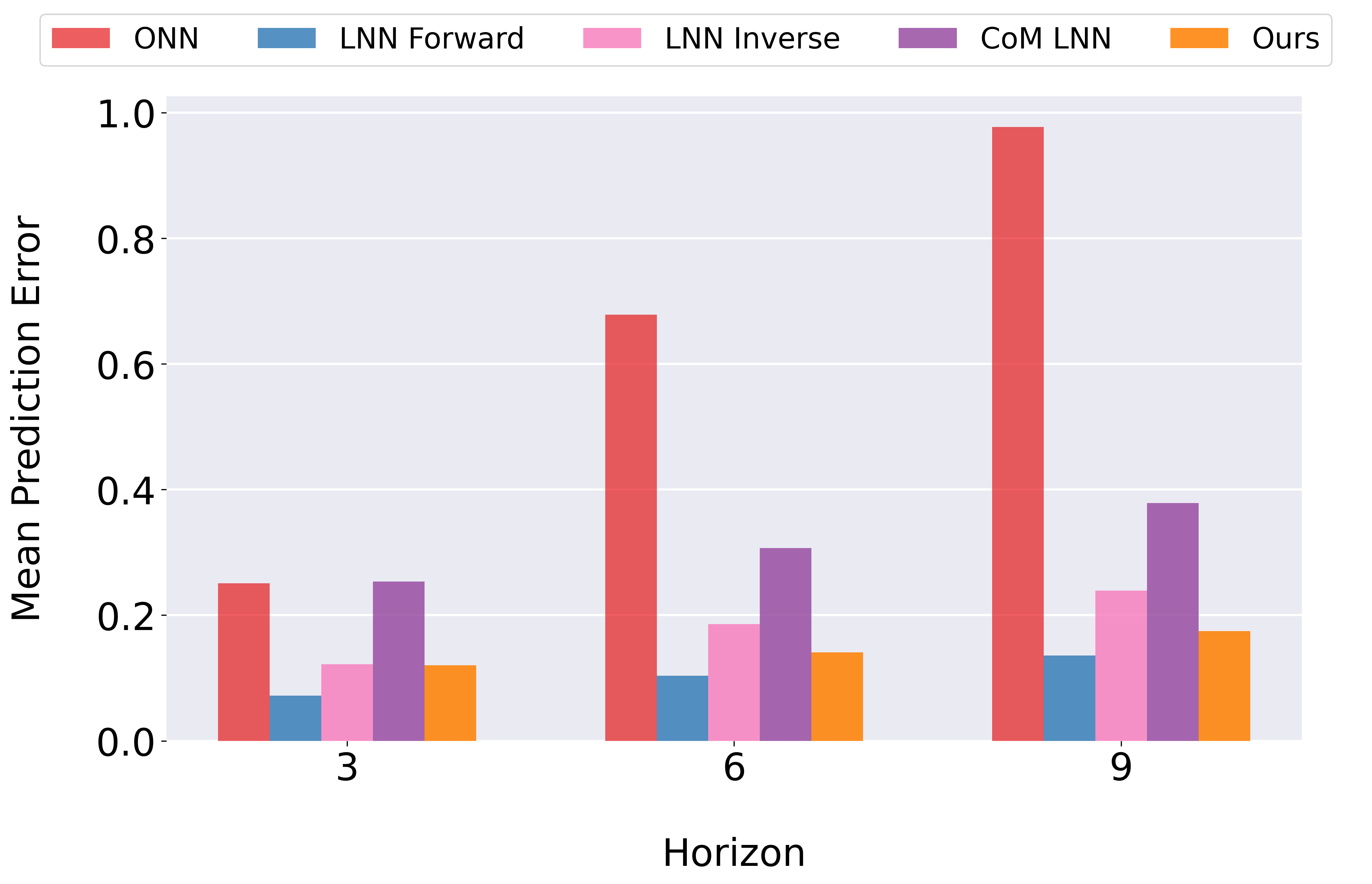}
    \caption{H-step prediction error of different models. Our approach maintains stable error across horizons, with significantly lower compounding error than ONN and CoM LNN.}
    \label{fig:prediction_error}
    \vspace{-10pt}
\end{figure}

\subsection{H-step Prediction Error}

We evaluate the prediction accuracy of the dynamics models over the horizon, which is critical for planning in MPC. Specifically, we measure the mean squared error between the predicted and actual simulation states (different for each method) over different rollout lengths.




Fig. \ref{fig:prediction_error} illustrates that the H-step prediction error across all models. ONN exhibits the highest prediction error, with severe compounding as the horizon increases. CoM LNN shows moderate improvements but still accumulates noticeable drift. In contrast, our method maintains stable error profiles even as the horizon increases, closely matching the performance of LNN Forward and LNN Inverse.

\subsection{Balancing Inference Speed and Performance}
\label{sc:mpc_speed_perf}

\begin{figure}[htp!]
    \centering
    \captionsetup{font=footnotesize}
    \includegraphics[width=0.8\linewidth]{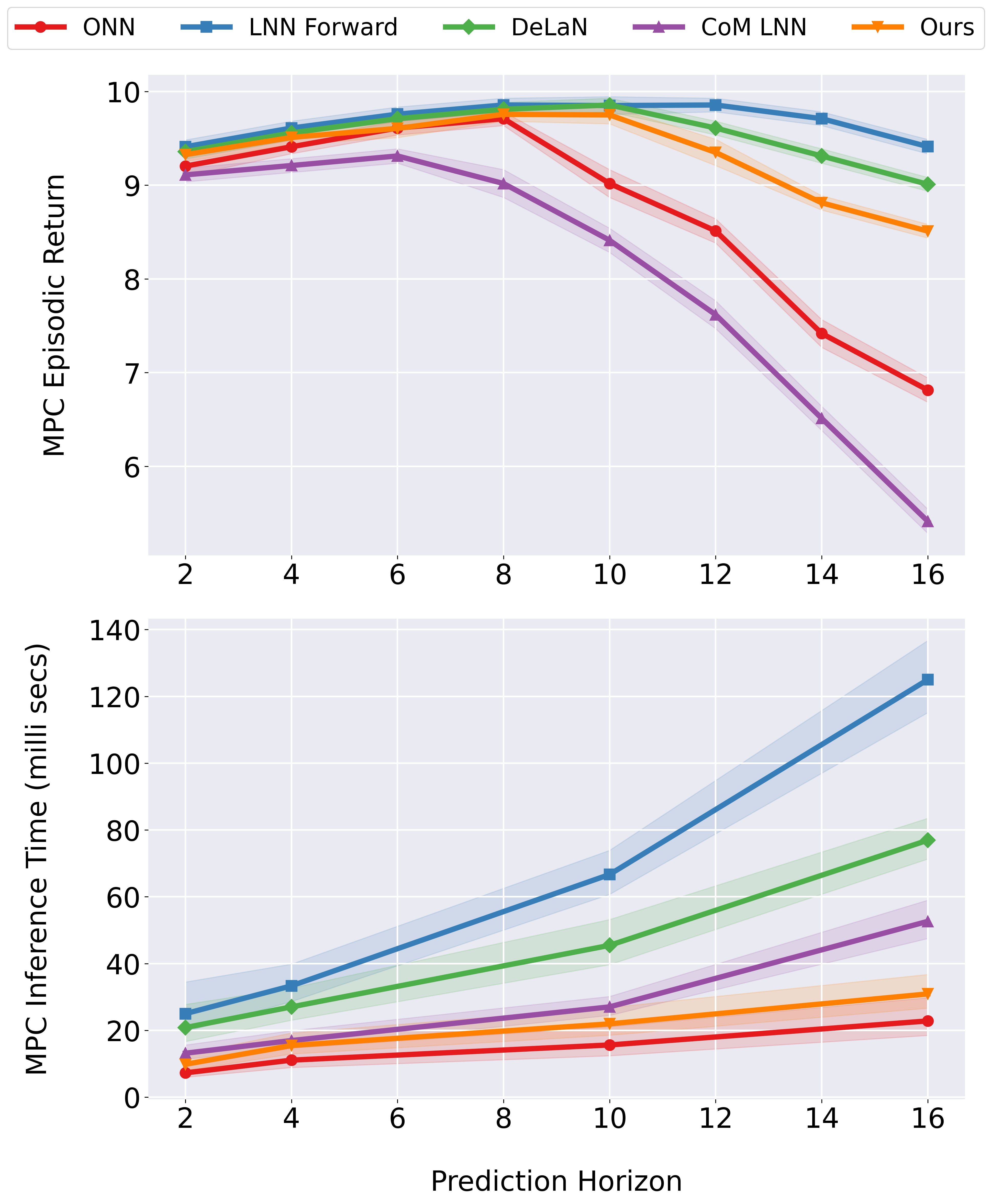}
    \caption{Trade-off between MPC inference time and control performance across planning horizons. Our inverse-dynamics-based planner achieves competitive returns with significantly lower inference latency.}
    \label{fig:combined_results}
\end{figure}

For deployment of MPC defined in Equation \eqref{eq:Forward_MPC} and \eqref{eq:Inverse_MPC}, a dynamics model must strike a delicate balance between inference speed and prediction accuracy. In this section, we evaluate all the methods for flat terrain in simulation across two key axes \footnote{While a full ablation of the warm-start component is beyond the scope of this work, MPPI is well known to benefit significantly from informed initialization \cite{mppi, piploco}; a cold-started sampler at H=8 with M=500 trajectories would require substantially more iterations to achieve the same solution quality, exceeding our real-time budget.} : (i) MPC inference time per control step, and (ii) Episodic return as a measure of control quality. Results are reported across increasing prediction horizons in Fig. \ref{fig:combined_results}.

\begin{table*}[htp!]
    \centering
    \captionsetup{font=footnotesize}
    \setlength{\arrayrulewidth}{1.5pt}
    \renewcommand{\arraystretch}{1.3}
    \caption{Performance comparison across terrains in simulation. Reported values are average episodic rewards ± standard deviation. DeLaN achieves the highest rewards, while our method closely matches with much lower compute cost.}
    \scriptsize
    \begin{tabular}{c|c|c|c|c|c|c}
    \hline
    \textbf{Method} & \textbf{Flat Ground} & \textbf{Slopes} & \textbf{Rough terrain} & \textbf{Stairs (12 cm)} & \textbf{Stairs (14 cm)} & \textbf{Stairs (16 cm)}\\
    \hline
    HIMLoco & $9.19 \pm 0.02$ & $9.18 \pm 0.05$ & $8.72 \pm 0.07$ & $7.32 \pm 0.24$ & $7.01 \pm 0.38$  & $6.31 \pm 0.58$ \\

    DeLaN (H=9) & $9.38 \pm 0.1$ & $9.38 \pm 0.11$ & $9.1 \pm 0.11$ & $8.33 \pm 0.29$ & $7.81 \pm 0.28$ & $7.31 \pm 0.31$ \\

    CoM LNN (H=5) & $9.05 \pm 0.26$ & $8.26 \pm 0.38$ & $7.98 \pm 0.62$  & $6.83 \pm 0.59$ & Failed & Failed \\

    ONN (H=9) & $9.19 \pm 0.07$ & $9.20 \pm 0.12$ & $8.80 \pm 0.21$ & $7.48 \pm 0.46$ & $7.21 \pm 0.48$ & $6.91 \pm 0.78$ \\

    Ours (H=9) & $9.23 \pm 0.17$ & $8.67 \pm 0.42$ & $8.64 \pm 0.42$ & $7.74 \pm 0.53$ & $7.51 \pm 0.58$ & $7.18 \pm 0.53$ \\
    \hline
    \end{tabular}
\label{tab:Perf_comp}
\vspace{-10pt}
\end{table*}

\textbf{ONN} achieves the lowest inference latency across all horizons. 
However, this comes at a clear cost: as the planning horizon grows beyond \(H = 8\), returns deteriorate significantly. The lack of physical priors leads to compounding errors and poor long-horizon reasoning. While ONN may be suitable for tasks with short lookahead, its utility in complex locomotion scenarios is limited.

\begin{figure*}[htbp!]
    \centering
    \captionsetup{font=footnotesize}
    \includegraphics[width=0.8\linewidth]{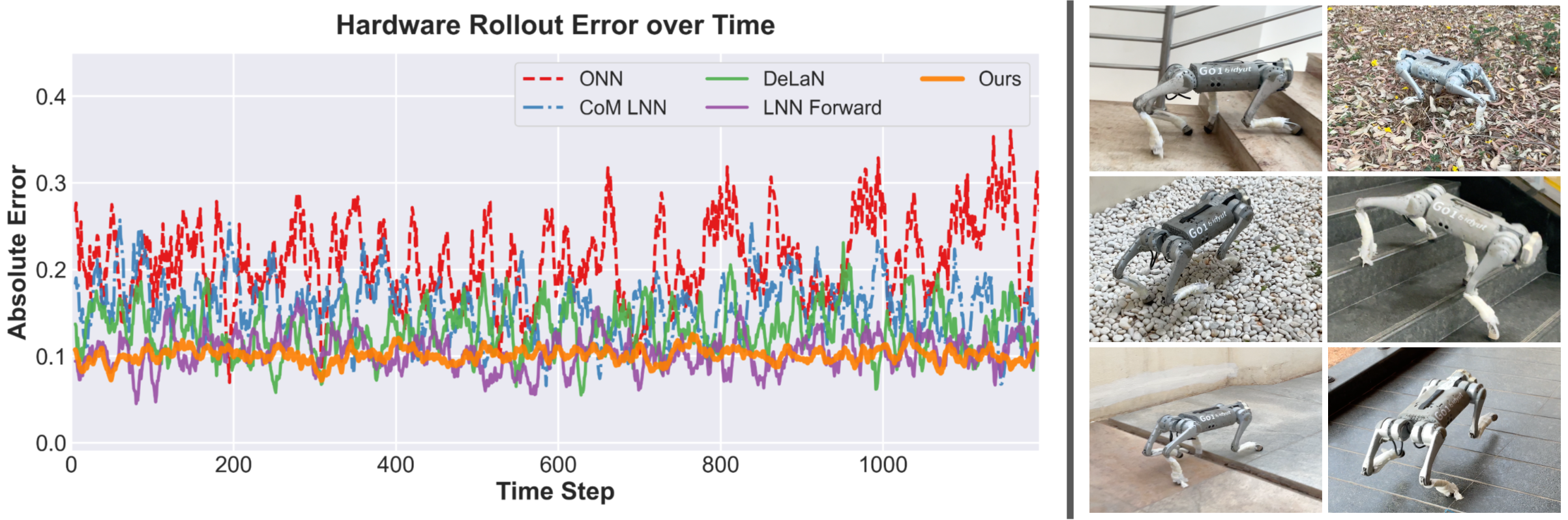}
    \caption{Hardware validation and absolute error tracking across 1200 time steps. Our method maintains significantly lower, more consistent rollout error than baselines (ONN, DeLaN, LNN variants) while navigating diverse terrain, including gravel, ramps, and stairs.}
    \label{fig:hardware}
    \vspace{-10pt}
\end{figure*}

\textbf{LNN Forward} provides a full-body Lagrangian model trained via forward dynamics and serves as a strong upper bound in terms of performance. It consistently achieves the highest returns across all horizon values. However, its use of full-matrix inversion during rollout leads to computational bottlenecks. While it can achieve near-optimal performance, it suffers from poor scalability, making it challenging to deploy under real-time constraints.

\textbf{DeLaN} exhibits strong performance across most horizons. Its structured formulation uses diagonalized mass matrices and energy-conserving properties to yield stable rollouts, even at high horizons. However, its inference time increases steeply with horizon length, reaching over \(80\) ms per control step at \(H=16\), making it unsuitable for high-frequency MPC control (e.g., \(>50\) Hz).


\textbf{CoM LNN}, which models only the CoM dynamics using Lagrangian structure, reduces computational cost relative to DeLaN. However, even this coarse-grained formulation incurs increasing latency with horizon due to the need for forward integration and mass matrix inversion. In terms of performance, CoM LNN performs reasonably well at short horizons but struggles at \(H > 10\), suggesting that the simplified dynamics limit its long-term predictive capabilities.

\textbf{Our method} replaces forward integration with inverse dynamics-based rollout, optimizing directly over joint positions \(q_{1:H}\). This key design choice allows for linear state propagation without matrix inversion, substantially reducing rollout cost. Our approach achieves an inference time of just \(30\) ms at \(H=16\), which is nearly \(4\times\) faster than LNN Forward and DeLaN. Despite this efficiency, it maintains competitive returns across all horizons, outperforming ONN and CoM LNN at \(H > 10\), and closely matching DeLaN and LNN Forward up to \(H = 12\). 
This demonstrates that our formulation not only offers computational benefits but also enables accurate long-horizon planning. Concretely, at $H=9$, our method achieves $98\%$ of DeLaN's episodic return on the most demanding stair terrain ($7.18$ vs. $7.31$) while running at approximately $4\times$ lower latency - making it the only planning-based method in our comparison capable of sustaining $50$ Hz control. This is a deliberate Pareto trade-off: a marginal reduction in peak reward in exchange for a latency profile compatible with real-time deployment on resource-constrained hardware.

To summarize, our MPC framework offers the best trade-off in the spectrum. It combines physical structure with efficient optimization, delivering strong returns and low latency even at large horizons. This makes it particularly suitable for real-world deployment where computational budgets are tight and long-term stability is essential.

\subsection{Multi-Terrain Performance}

We evaluate the proposed method's robustness and generalization across diverse terrains, both in simulation and on hardware. Our goal is to assess whether the inverse-dynamics-based controller retains long-horizon planning capability and inference efficiency in real-world settings.

\subsubsection{Simulation Results} Table \ref{tab:Perf_comp} presents average rewards on flat ground, slopes, rough terrain, and stairs (of increasing height), comparing our method against baselines. Among the planning-based controllers, DeLaN achieves the highest rewards, benefiting from its structured formulation and stability under forward rollouts. However, this performance comes at a significant computational cost.

Our method performs competitively across all terrains, closely matching DeLaN on flat ground and outperforming CoM LNN and ONN on stairs. Despite using inverse dynamics for rollout, it maintains sufficient predictive consistency to enable effective long-horizon planning. Unlike ONN, which lacks physical priors and exhibits degraded performance in complex terrains, our model benefits from structured regularization without incurring high inference latency.
We also include HIMLoco \cite{himloco} as a non-planning, end-to-end RL baseline. While it performs reasonably on all terrains, it lacks interpretability or benefits of planning.


\subsubsection{Hardware Validation} 
We deploy our method on a Unitree Go1 robot and test on unstructured outdoor environments including loose gravel, ramps, flat pavement, and staircases. Fig. \ref{fig:hardware} shows representative deployment scenarios alongside the absolute hardware rollout error over 1200 time steps. Our method maintains consistently lower, more stable error (averaging $\sim$0.1) than the ONN, and other LNN variants, which exhibit significantly higher variance and drift. 

On hardware, the controller remains reactive to terrain irregularities such as foot slippage or variable stair heights, thanks to its 50Hz on-device rollout speed and predictive consistency. In stair climbing, it recovers from foot perturbations without the use of pre-programmed heuristics. This robustness demonstrates successful sim-to-real transfer.




\section{Conclusion}

We presented a control framework that integrates Lagrangian Neural Networks with model predictive control and reinforcement learning for quadrupedal locomotion. By leveraging inverse dynamics, our method improves sample efficiency during training and reduces inference overhead during deployment. Simulation benchmarks show favorable trade-offs between prediction accuracy and runtime performance, and hardware tests on a Unitree Go1 robot confirm feasibility across diverse terrains with minimal tuning. Future work includes refining inverse dynamics formulations, incorporating contact dynamics, and extending the approach to more complex locomotion tasks involving perception.

\bibliographystyle{ieeetr}
\footnotesize
\bibliography{references}






\end{document}